\pdfoutput=1

\documentclass[11pt,a4paper]{article}
\usepackage[hyperref]{acl2020}
\usepackage{times}
\usepackage{latexsym}

\newcommand{\argmin}{\mathop{\rm argmin}\limits}

\usepackage{microtype}

\aclfinalcopy %
\def\aclpaperid{1112} %

\usepackage{booktabs}
\usepackage{setspace}
\usepackage{graphicx}

\title{Evaluating Dialogue Generation Systems via Response Selection}

\author{
Shiki\,Sato$^{1}$\hspace{1em}
Reina\,Akama$^{1,2}$\hspace{1em}
Hiroki\,Ouchi$^{2,1}$\hspace{1em}
Jun\,Suzuki$^{1,2}$\hspace{1em}
Kentaro\,Inui$^{1,2}$\\[2pt]
$^{1}$Tohoku University\hspace{1em}
$^{2}$RIKEN\hspace{1em}
\\
\texttt{\{shiki.sato,reina.a,jun.suzuki,inui\}@ecei.tohoku.ac.jp}
\\
\texttt{hiroki.ouchi@riken.jp}
}
\date{}

\begin{document}
\maketitle
\begin{abstract}
Existing automatic evaluation metrics for open-domain dialogue response generation systems correlate poorly with human evaluation. 
We focus on evaluating response generation systems via response selection.
To evaluate systems properly via response selection, we propose the method to construct response selection test sets with well-chosen false candidates.
Specifically, we propose to construct test sets filtering out some types of false candidates: (i) those unrelated to the ground-truth response and (ii) those acceptable as appropriate responses.
Through experiments, we demonstrate that evaluating systems via response selection with the test sets developed by our method correlates more strongly with human evaluation, compared with widely used automatic evaluation metrics such as BLEU.
\end{abstract}

\section{Introduction}
\label{sec:introduction}
Automatic evaluation for open-domain dialogue generation systems has a potential for driving their research and development because of its high reproducibility and low cost.
However, existing automatic evaluation metrics, such as BLEU \cite{papineni-etal-2002-bleu}, correlate poorly with human evaluation \cite{liu-etal-2016-evaluate}.
This poor correlation arises from a nature of dialogue, that is, there are many acceptable responses to an input context, known as the one-to-many problem \cite{zhao-etal-2017-learning}.

To tackle this problematic issue, we focus on evaluating response generation systems via response selection.
In this task, systems select an appropriate response for a given context from a set of response candidates.
Each candidate has the label that indicates whether the candidate is appropriate for the given context.
Traditionally, response selection has been used to evaluate retrieval-based dialogue systems \cite{lowe-etal-2015-ubuntu,wu-etal-2017-sequential}.
Here, we consider applying response selection to driving the research for dialogue generation systems.
Specifically, we consider using response selection to pick out promising systems that should be evaluated more precisely by humans among a lot of candidate systems.
We assume that response selection is a valid option for such a preliminary evaluation on the basis of the following assumption: systems that can generate appropriate responses can also select appropriate responses.
One advantage of evaluating generation systems via response selection is that it can remedy the one-to-many problem, because we do not have to consider the appropriate responses that are not included in sets of response candidates.
Another advantage is that it enables a simple and clear comparison between systems in accuracy.

\begin{figure}[t]
\begin{center}
\includegraphics[width=75mm]{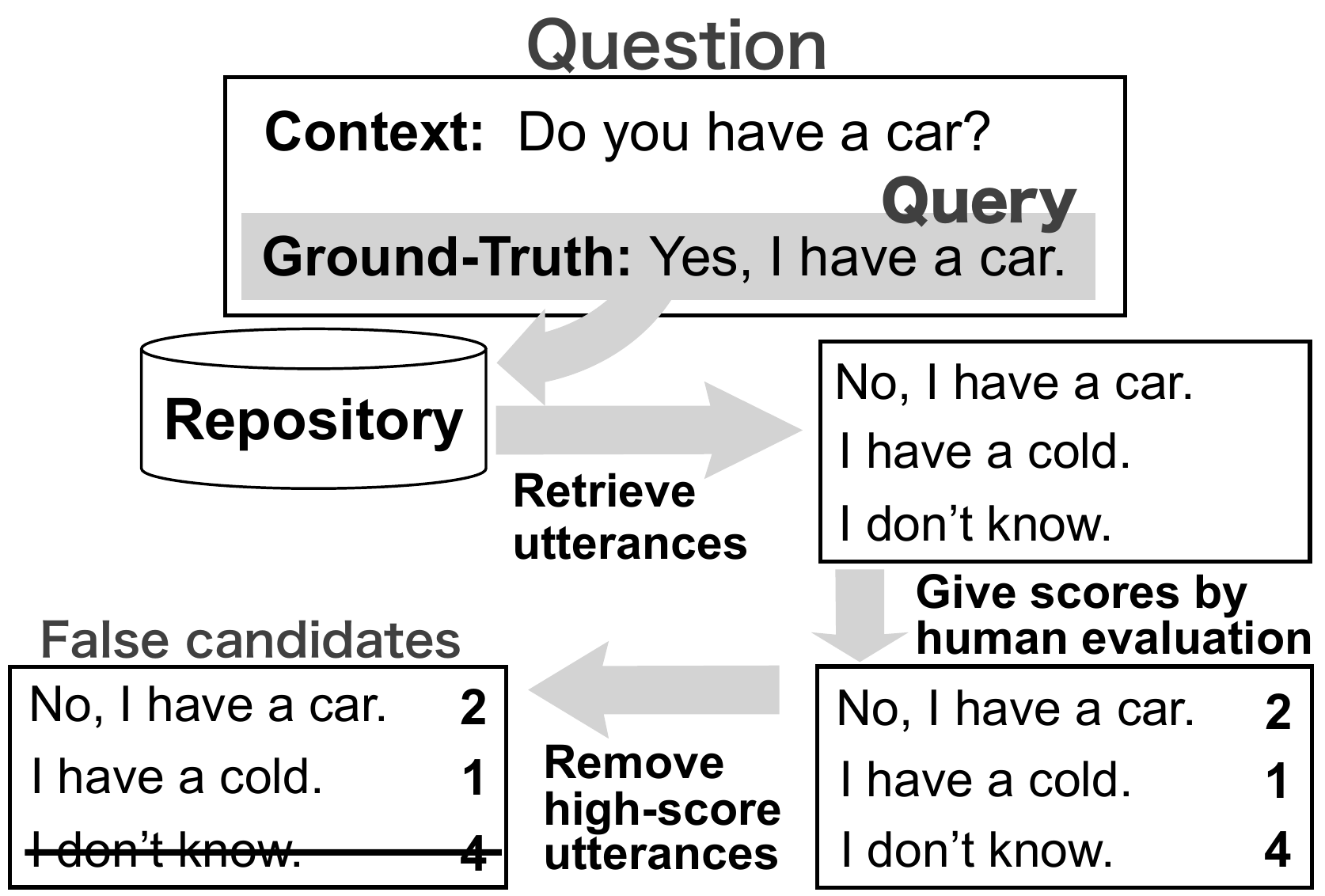}
\end{center}
\caption{
Overview of the construction method of our test set.
First, we retrieve only utterances related to the ground-truth response from a repository.
Then, we remove acceptable utterances by human evaluation.}
\label{fig:figure1}
\end{figure}

Generally, false response candidates are randomly sampled from a repository\cite{lowe-etal-2015-ubuntu,gunasekara-etal-2019-dstc7}, which causes two problems: (i) unrelated false candidates and (ii) acceptable utterances as false.
The first problem is that randomly sampled false candidates are often too far from ground-truth responses.
Consider the case where for a given context ``Do you have a car?", a response candidate ``I play tennis." is randomly sampled.
Systems can easily recognize this candidate as a false one because there are no related content words between them.
Such excessive easiness is not preferable because the performance gap between good and inferior systems tends to be small.
The second problem is that there is no guarantee that randomly sampled candidates are always unacceptable ones.
For example, ``I don't know." is often sampled as a false response because this phrase often occurs in open-domain dialogues.
This phrase can be regarded as acceptable for various contexts.
These two problems make general response selection test sets unreliable.

In this work, we propose a method to construct response selection test sets with well-chosen false candidates (Figure \ref{fig:figure1}).
First, we retrieve only utterances related to the ground-truth response.
Then we remove acceptable utterances by human evaluation.
Through experiments, we demonstrate that automatic evaluation using the test set developed by our method correlates more strongly with human evaluation, compared with widely used automatic evaluation metrics such as BLEU.
Our empirical results indicate that response selection with well-chosen false candidates can be a valid option for evaluating response generation systems.
We will release the test set used in the experiments.\footnote{The test set is available at \url{https://github.com/cl-tohoku/eval-via-selection}.}

\section{Related Work}

\paragraph{Automatic evaluation metrics}
Various metrics have been proposed for automatic evaluation of dialogue systems, such as BLEU, METEOR \cite{banerjee-lavie-2005-meteor}, ROUGE \cite{lin-2004-rouge},  Greedy Matching \cite{rus-lintean-2012-comparison}, and Vector Extrema \cite{forgues2014bootstrapping}.
These metrics evaluate the quality of the responses generated by systems.
However, this is challenging due to the one-to-many problem.
For example, ADEM, a metric proposed by \cite{lowe-etal-2017-towards}, is easily fooled by adversarial examples (responses) \cite{Sai2019ReevaluatingAA}.
To remedy one-to-many problem, we focus on evaluating systems via response selection.

\paragraph{Response selection test sets with human labels}
One popular test set for response selection is Douban Conversation Corpus in Chinese \cite{wu-etal-2017-sequential}.
In this test set, each response candidate has a manually annotated label that indicates whether or not the candidate is appropriate for the given context.
Although this test set is similar to ours, there are some differences between the purposes and procedure of test set designs.
The purpose of creating their test set is to simulate and evaluate retrieval-based dialogue systems.
Thus, all the candidates in this corpus are retrieved by using the context as queries, as retrieval-based systems do.
In this paper, we develop an English response selection test set with human labels to evaluate dialogue generation systems.
One of the salient differences from Douban Conversation Corpus is the procedure of retrieving false candidates.
We retrieve false candidates using the ground-truth responses.
By this method, we can more certainly collect false candidates that are related to ground-truth responses and facilitate error analysis as described in Section \ref{subsec:Discussion}.

\section{Test Set Construction}\label{sec:TestsetConstruction}

\subsection{Construction Method}
\label{subsec:method}

For each context $c$ and ground-truth response $r^{\rm true}$, we construct a set of false response candidates $r^{\rm false} \in \mathcal{R}^{\rm false}$ by retrieving utterances from an utterance repository $u \in \mathcal{U}$.
As we mentioned in Section~\ref{sec:introduction}, we want to filter out some types of utterance: (i) those unrelated to the ground-truth response and (ii) those acceptable as appropriate responses.
We filter out such utterances as follows:
\begin{enumerate}
    \setlength{\parskip}{0cm}
    \setlength{\itemsep}{0cm}
    \item Retrieve $M$ utterances, $\{ u_1, \cdots, u_M \}$, related to the ground-truth response $r^{\rm true}$ from the utterance repository $\mathcal{U}$.
    \item Remove acceptable ones from the retrieved utterances by human evaluation.
\end{enumerate}

\paragraph{1. Retrieve utterances related to the ground-truth response}
We assume that utterances related to the ground-truth response share some similar content words between them.
Here, we retrieve the related utterances on the basis of the similarities of the content words.
This process makes it difficult for systems to distinguish between ground-truth and false candidates only by comparing the content words.

\paragraph{2. Remove acceptable utterances}
Coincidentally, some of the retrieved utterances may be acceptable as an appropriate response.
To remove such utterances, we ask human annotators to evaluate each retrieved utterance.
Specifically, we instruct five annotators (per candidate) to score each retrieved candidate in a five-point scale from $1$ to $5$.
A score of $5$ means that the utterance can clearly be regarded as an appropriate response for the given context, whereas a score of $1$ means that it cannot be regarded as an appropriate one at all.
In addition to the scores, we also instruct annotators to give a score of $0$ to ungrammatical utterances.
We remove the utterances that were given a score of $3$ or higher by three or more annotators because these utterances with a high score can be acceptable.
In addition, we remove the utterances that were given a score of $0$ by three or more annotators because these are likely to be ungrammatical ones.
We also instruct annotators to score ground-truth responses, combining them with retrieved candidates.
We remove the questions if the score of the ground-truth response is low, i.e., three or more annotators gave a score of $3$ or lower.
This is intended to ensure that ground-truth responses are certainly appropriate for the given context.

\subsection{Overview of Constructed Test Set}

\paragraph{Settings of test set construction}
We retrieve $10$ utterances (per question) from the repository and remove acceptable ones following the method described in Section \ref{subsec:method}.
We use crowdsourcing\footnote{https://www.mturk.com/} to score the retrieved utterances.
After removing acceptable utterances from retrieved ones, there are some questions that have $6$ or more available false candidates.
From these questions, we develop new questions with the same context but different  candidates (both ground-truth responses and false candidates).
We use removed (acceptable) utterances for the ground-truth responses of new questions.

We use the dialogue data from DailyDialog \cite{li-etal-2017-dailydialog} to construct the test set. We extract the four beginning turns of each dialogue sample from DailyDialog, regarding the fourth utterance as the ground-truth response.
We extract the utterances of OpenSubtitles2018 \cite{lison-etal-2018-opensubtitles2018} to construct the repository used to retrieve false candidates.
Note that the repository does not contain the utterances in the dialogue data used to train response generation systems in Section \ref{subsec:experimentalprocedure}.

\paragraph{Statistics of our test set}
We developed the test set that consists of $1,019$ questions with $4$ candidates ($1$ ground-truth + $3$ false candidates).

Table \ref{tab:statistics} shows the basic statistics of our test set.
The Fleiss' Kappa \cite{Fleiss71} of the annotators' scoring in the six scale is $0.22$.\footnote{We calculated Fleiss’ Kappa based on the scale of the scores as categorical.}
Note that if we regard the scoring as binary classification (scores higher than $3$ are regarded as appropriate responses, and the others not), the Fleiss' Kappa of the scoring is $0.63$, which is higher than Douban Conversation Corpus ($0.41$).

\begin{table}[t]
\centering
\begin{tabular}{lr} \toprule
Total questions              & 1,019 \\
Candidates per question      & 4     \\
Context turns per question   & 3     \\
Kappa of the scoring (six classes) & 0.22 \\
Kappa of the scoring (two classes) & 0.63 \\ \bottomrule
\end{tabular}
\caption{Basic statistics of our test set}
\label{tab:statistics}
\end{table}

\paragraph{Example of our test set}
Table \ref{tab:example} shows an example of our test set.
All the false response candidates share the same content word ``focus" related to the topic ``camera".

\paragraph{Preliminary experiments}
We conducted a simple experiment to investigate whether or not a system that takes only content words into account can recognize false response candidates in our test set.
For the model, we used the TF-IDF model \cite{lowe-etal-2015-ubuntu}, which simply compares between content words of a given context and each candidate.
As a result, the accuracy was $0.461$.
For a comparison, we also replaced all the false candidates in our test set with randomly sampled utterances.
The accuracy of the same TF-IDF model increased to $0.671$.
These results indicates that it is difficult to recognize false candidates in our test set only by comparing content words.

\begin{table}[t]
    \centering
    \small
    \begin{tabular}{rp{40ex}}
        \toprule
        \multicolumn{2}{l}{\textbf{Context:}} \\
        A: & Excuse me. Could you please take a picture of us with this {\bf camera?} \\
        \rule{0pt}{2.5ex}
        B: & Sure. Which button do I press to shoot? \\
        \rule{0pt}{2.5ex}
        A: & This one. \\
        \midrule
        \multicolumn{2}{l}{\textbf{Candidates:}} \\
        1. & Could he not \textbf{focus} on that? \\
        2. & But I do have ninja \textbf{focus}. \\
        3. & Do not lose your \textbf{focus}! \\
        4. & Do I have to \textbf{focus} it? [Ground-truth] \\
        \bottomrule
    \end{tabular}
    \caption{Example of our test set. All three false candidates contain the content word ``focus", which is related to the context (topic). }
    \label{tab:example}
\end{table}

\section{Experiments}
We test whether the automatic evaluation of response generation systems on our test set correlates with human evaluation.

\subsection{Experimental Procedure}
\label{subsec:experimentalprocedure}
We train multiple response generation systems and rank them on the basis of human and automatic evaluation scores.
By comparing between the system ranking by human scores and the ranking by each automatic score, we verify the correlations.
\subsubsection{Response Generation Models}

We train $10$ different response generation systems to be ranked in the experiments.
Their architectures are ones of Seq2Seq with GRU \cite{cho-etal-2014-learning}, Seq2Seq with LSTM \cite{lstm}, or Transformer \cite{transformer}.
Some systems have same architecture, but different hyper-parameters.\footnote{We describe the model settings in Appendix \ref{sec:DetailedExperimentalSettings}.}

We train the models on OpenSubtitles2018.
The training data consists of 5M samples and the validation data consists of 0.05M samples, each of which is four-turns dialogue.

\subsubsection{Evaluation Procedure}

\paragraph{Ground-truth system ranking by human scores}
The trained systems generate a response $r^{\rm gen}$ for each input context $c \in \mathcal{C}$.
Then, five human annotators (per response) score each generated response $r^{\rm gen}$ in a five-point scale from 1 to 5.
A score of 5 means that the response can clearly be regarded as an appropriate response for the given context, whereas a score of 1 means that it cannot be regarded as an appropriate one at all.
As a result, we obtain five scores, $\{s_1, s_2, \cdots, s_5\}$, for each response $r^{\rm gen}$ and average them: $s^{\rm mean} = {\rm mean}(s_1, s_2, \cdots, s_5)$.
We also average $s^{\rm mean}$ across all the questions in the test set and yield the final score $s^{\rm final}$ for each system.
Based on this score, we make a ranking of the systems and regard it as the ground-truth ranking.
We developed the test set that consists of 1,019 questions.
However, it is too costly to evaluate all the 10 systems’ responses for 1,019 questions by humans.
Thus we give the context of $56$ randomly sampled questions from our test set to the 10 systems as inputs $\mathcal{C}$.

\paragraph{System ranking by response selection accuracy}
We rank the systems by response selection accuracy with well-chosen false candidates (CHOSEN).
The trained response generation systems compute the softmax cross-entropy loss $\ell_r$ for each response candidate $r \in \mathcal{R}$.
We regard the candidate with the lowest loss as the system's selection: $\hat{r} = \argmin_{r \in \mathcal{R}} \ell_r$.
From the predictions, we calculate accuracy and make a ranking of the systems based on the accuracy.
For comparison, we also make a ranking by response selection accuracy with randomly sampled false candidates (RANDOM).\footnote{We compute the coefficient of RANDOM by averaging the coefficients of different 100 trials.}
We compute the accuracy of CHOSEN and RANDOM using all $1,019$ questions from our test set.

\paragraph{System ranking by other evaluation metrics}
For comparison, we also make rankings of the systems by three existing automatic evaluation metrics: BLEU, METEOR, and ROUGE-L.
First, the trained systems generate a response for each input context.
Then we compute the scores comparing generated responses and the ground-truth responses.
These scores can be computed without false candidates.
Thus we compute them using all $7,393$ available four-turns dialogue samples from DailyDialog, regarding the fourth utterances as the ground-truth responses.

\begin{table}[t]
\centering
\begin{tabular}{@{}lcc@{}}
\toprule
Metrics & Spearman & p-value \\ \midrule
BLEU-1  & \hspace*{0.2mm}$-$0.36    & 0.30    \\
BLEU-2  & \hspace*{4.2mm}0.085    & 0.82    \\
METEOR  & \hspace*{4.2mm}0.073    & 0.84    \\
ROUGE-L   & \hspace*{2.7mm}0.35     & 0.33    \\ \midrule %
RANDOM  & \hspace*{2.7mm}0.43     & -    \\
{\bf CHOSEN}  &\hspace*{1.9mm} {\bf 0.48}     & {\bf 0.19}    \\ \midrule %
HUMAN  & \hspace*{2.7mm}0.87     & \hspace*{3.1mm}0.0038  \\ \bottomrule
\end{tabular}
\caption{Correlations between the ground-truth system ranking and the rankings by automatic evaluation.}
\label{tab:correlation}
\end{table}

\subsection{Results}
We compare the rankings by Spearman's rank correlation coefficients, shown in Table \ref{tab:correlation}.
First, we yielded the human upper bound.
we evaluated the correlation between the rankings made by different annotators (HUMAN).
We randomly divided human evaluation into two groups and made two rankings.
The correlation coefficient between the two rankings was $0.87$.
Second, we found that the rankings made using existing automatic evaluation metrics correlate poorly with ground-truth ranking.
BLEU, often used to evaluate generation systems, does not correlate with human evaluation at all.
One exception is ROUGE-L.
However, its correlation coefficient is lower than $0.4$, which means reasonable correlation.
Third, we found that the ranking made by using our test set reasonably correlates with the ground-truth ranking compared with other metrics, and the correlation coefficient (CHOSEN) is higher than $0.4$.

\begin{figure}[t]
\begin{center}
\includegraphics[width=75mm]{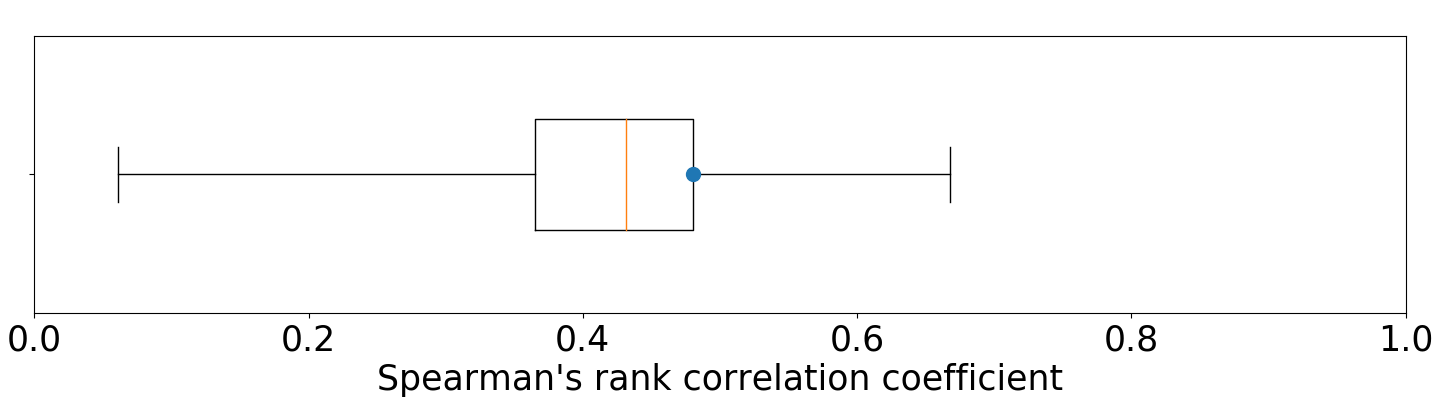}
\end{center}
\caption{Box plot of Spearman's rank correlation coefficients between the ground-truth ranking and the rankings by RANDOM. A dot in blue indicates the correlation coefficient of CHOSEN.}
\label{fig:boxplot}
\end{figure}

\begin{table}[t]
    \centering
    \small
    \begin{tabular}{rp{40ex}}
        \toprule
        \multicolumn{2}{l}{\textbf{Context:}} \\
        A: & Peter, enough with your computer games. Go do your homework now.\\
        \rule{0pt}{2.5ex}
        B: & Can't I play more?   \\
        \rule{0pt}{2.5ex}
        A: & No! Stop playing computer games! \\
        \midrule
        \multicolumn{2}{l}{\textbf{Candidates:}} \\
        \multicolumn{2}{l}{Ground-Truth: Mom, I'll be finished soon.}  \\
        \multicolumn{2}{l}{\quad RANDOM: Thats the problem with small towns.} \\
        \multicolumn{2}{l}{\quad\,\, CHOSEN: You are to be finished very soon.} \\
        \bottomrule
    \end{tabular}
    \caption{Examples of a randomly sampled and well-chosen candidates.}
    \label{tab:comparisoncandidates}
\end{table}

\subsection{Discussion} \label{subsec:Discussion}
\paragraph{Instability of evaluation with random sampling}
The correlation coefficient of the ranking by response selection with randomly sampled false candidates (RANDOM) is higher than that of BLEU and slightly lower than that of CHOSEN.
However, a serious problem has been observed: the instability.
We make $100$ test sets, each of which consists of different false candidates by random sampling with different seeds.
For each test set, we make a system ranking and compute its coefficient.
Figure \ref{fig:boxplot} shows the box plot of the Spearman's rank correlation coefficients of the trials.
The range of the coefficients is very wide (0.06-0.67).
This result means that the quality of evaluation with randomly sampled false candidates strongly depends on the sampled candidates, which is the uncontrollable factor stemming from the randomness.

\paragraph{Interpretable error analysis}
Our automatic evaluation with well-chosen false candidates brings another benefit: the interpretable error analysis.
Table \ref{tab:comparisoncandidates} shows an example of a question of our test set.
The well-chosen false candidate (CHOSEN) is similar to the ground-truth response.
However, the grammatical subject of the CHOSEN sentence is ``You", which completely mismatches the context.
Thus if systems select this false candidate, they may lack the ability to determine correctly the subject of sentences.
In this way, our test set enables us to analyze systems' predictions from various meaningful perspectives.
As a case study, we design a set of error labels, each of which indicates why the false candidate is false, and assign them to $50$ false candidates in our test set.
We succeed in assigning the labels to $22$ out of $50$ candidates.\footnote{We show some of the examples in Appendix \ref{sec:labels}.}

\paragraph{Limitation}
Our test set is designed to evaluate open-domain dialogue generation systems.
Thus, it is not suitable for evaluating other types of dialogue system such as task-oriented ones.
By contrast, existing automatic evaluation metrics, such as BLEU, do not have this type of restriction.

\section{Conclusion}
In this paper, we focused on evaluating response generation systems via response selection.
To evaluate systems properly via response selection, we proposed the method to construct response selection test sets with well-chosen false candidates.
Specifically, we proposed to construct test sets filtering out some types of false candidates: (i) those unrelated to the ground-truth response and (ii) those acceptable as appropriate responses.
We demonstrated that evaluating systems via response selection with the test sets developed by our method correlates more strongly with human evaluation, compared with that of widely used metrics such as BLEU.

In the future, we will provide labels that indicate ``Why this candidate is false" for false candidates in our test set, so that one can easily detect weak points of systems through error analysis.

\section*{Acknowledgments}
This work was partially supported by JSPS KAKENHI Grant Number JP19H04162.
We would like to thank the laboratory members who gave us advice and all reviewers of this work for their insightful comments.

\clearpage

\bibliography{acl2020}
\bibliographystyle{acl_natbib}

\clearpage
\appendix

\section{Methods to Retrieve False Candidates}
\label{sec:constructionsettings}
To make false candidates in each pool diverse, we use two retrieval methods: lexical retrieval and embedding-based retrieval.
We use Lucene\footnote{https://lucene.apache.org/} for lexical retrieval, and cosine similarity of sentence vectors for embedding-based retrieval.
Sentence vectors are SIF \cite{Arora2017} weighted average of ELMo word vectors \cite{peters-etal-2018-deep}.

\section{Detailed Model Settings in the Experiments}\label{sec:DetailedExperimentalSettings}
We trained $10$ different response generation systems to be ranked in the experiments. 
We trained them with different architectures or settings. 
The common settings for the model training are shown in Table \ref{tab:model1} and the hyper-parameters of each the models are shown in Table \ref{tab:model2}.

\begin{table}[h]
\small
\tabcolsep 14pt
\centering
\begin{tabular}{p{8em}r}
\toprule
Vocab size      & 16,000        \\
Batch size      & 6,000 tokens  \\
Loss            & cross entropy \\
Learning rate   & 1e-4 (fixed)  \\
Optimizer       & Adam          \\ 
\bottomrule
\end{tabular}
\caption{Common settings for the model training in the experiments.}
\label{tab:model1}
\end{table}

\begin{table}[h]
\small
\tabcolsep 2.5pt
\centering
\begin{tabular}{c cccc}
\toprule
No. &
Architecture &
\begin{tabular}{c}Enc/Dec\\layers\end{tabular}& 
\begin{tabular}{c}Enc/Dec\\embed dim\end{tabular}& 
\begin{tabular}{c}Enc/Dec\\hidden dim\end{tabular}      \\ \midrule %
1      & GRU          & 1 / 1                  & 256 / 256                 & 256 / 256                       \\
2      & GRU          & 1 / 1                  & 512 / 512                 & 512 / 512                       \\
3      & GRU          & 2 / 2                  & 256 / 256                 & 256 / 256                       \\
4      & GRU          & 2 / 2                  & 512 / 512                 & 512 / 512                       \\
5      & LSTM         & 1 / 1                  & 256 / 256                 & 256 / 256                       \\
6      & LSTM         & 1 / 1                  & 512 / 512                 & 512 / 512                       \\
7      & LSTM         & 2 / 2                  & 512 / 512                 & 512 / 512                       \\
\midrule %
No. &
Architecture &
\begin{tabular}{c}Enc/Dec\\layers\end{tabular}& 
\begin{tabular}{c}Enc/Dec\\embed dim\end{tabular}& 
\begin{tabular}{c}Enc/Dec\\attention heads\end{tabular}      \\ \midrule %
8      & Transformer  & 2 / 2                  & 256 / 256                 & 4 / 4                           \\
9      & Transformer  & 2 / 2                  & 512 / 512                 & 4 / 4                           \\
10     & Transformer  & 4 / 4                  & 256 / 256                 & 4 / 4                           \\
\bottomrule
\end{tabular}
\caption{Hyper-parameters of each model in the experiments.}
\label{tab:model2}
\end{table}

\section{Labels for False Candidates}
\label{sec:labels}
As a case study, we designed a set of error labels, each of which indicates why the false candidate is false.
To confirm whether we can assign the labels to the false candidates collected by our test set construction method, We assigned the labels to $50$ false candidates from our test set.
We could eventually assign the labels to $22$ candidates.
The types of our error labels and the breakdown are listed in Table \ref{tab:breakdown}.
The examples of false candidates (CHOSEN) corresponded to the error labels are shown in Table \ref{tab:comparisoncandidates} (for labeled ``Responses that have wrong subjects''), Table \ref{tab:inconsistent}, Table \ref{tab:insufficient}, and Table \ref{tab:tense}.

\begin{table}[h]
\centering
\small
\begin{tabular}{lc} 
\toprule
Error label & Count \\
\midrule
Inconsistent responses with the context              & 8 \\
Responses that have insufficient information      & 4     \\
Responses that have wrong subjects   & 9     \\
Responses with wrong tense & 1 \\ 
\bottomrule
\end{tabular}
\caption{Error labels and the breakdown of the the assigned labels.}
\label{tab:breakdown}
\end{table}

\begin{table}[!h]
    \centering
    \small
    \begin{tabular}{rp{40ex}}
        \toprule
        \multicolumn{2}{l}{\textbf{Context:}} \\
        A: & 911 emergency. What is the problem?\\
        \rule{0pt}{2.5ex}
        B: & I would like to report a break-in.   \\
        \rule{0pt}{2.5ex}
        A: & When was this break-in?  \\
        \midrule
        \multicolumn{2}{l}{\textbf{Candidates:}} \\
        \multicolumn{2}{l}{Ground-Truth: I believe it happened last night.}  \\
        \multicolumn{2}{l}{\quad\,\, CHOSEN: I thought that would happen last night. } \\
        \bottomrule
    \end{tabular}
    \caption{Example of a false candidate labeled ``Inconsistent responses with the context.''}
    \label{tab:inconsistent}
\end{table}

\begin{table}[!h]
    \centering
    \small
    \begin{tabular}{rp{40ex}}
        \toprule
        \multicolumn{2}{l}{\textbf{Context:}} \\
        A: & What’s the matter with you, Paul?\\
        \rule{0pt}{2.5ex}
        B: & I’m not feeling well. I think I’m having a cold.   \\
        \rule{0pt}{2.5ex}
        A: & Looks like it. You need to drink a lot of water and take a good rest. \\
        \midrule
        \multicolumn{2}{l}{\textbf{Candidates:}} \\
        \multicolumn{2}{l}{Ground-Truth: Yeah, I will.}  \\
        \multicolumn{2}{l}{\quad\,\, CHOSEN: Yeah, yeah, yeah, I...} \\
        \bottomrule
    \end{tabular}
    \caption{Example of a false candidate labeled ``Responses that have insufficient information.''}
    \label{tab:insufficient}
\end{table}

\begin{table}[!h]
    \centering
    \small
    \begin{tabular}{rp{40ex}}
        \toprule
        \multicolumn{2}{l}{\textbf{Context:}} \\
        A: & Hi, charlie, are you busy this evening?\\
        \rule{0pt}{2.5ex}
        B: & Sorry, I'm afraid that  I've got plans tonight.   \\
        \rule{0pt}{2.5ex}
        A: & What are you doing? \\
        \midrule
        \multicolumn{2}{l}{\textbf{Candidates:}} \\
        \multicolumn{2}{l}{Ground-Truth: I'm going to my parents'house}\\
        \multicolumn{2}{l}{\hspace{6em} for my father's birthday.}\\
        \multicolumn{2}{l}{\quad\,\, CHOSEN: We were at my sister’s house for my}\\
        \multicolumn{2}{l}{\hspace{6em} nephew’s birthday by 2 p.m.}\\
        \bottomrule
    \end{tabular}
    \caption{Example of a false candidate labeled ``Responses with wrong tense.''}
    \label{tab:tense}
\end{table}

\end{document}